%% file: main.tex
\documentclass{article}

\usepackage{PRIMEarxiv}
\usepackage{tabto}
\usepackage[utf8]{inputenc} 
\usepackage[T1]{fontenc}    
\usepackage{hyperref}       
\usepackage{url}            
\usepackage{booktabs}       
\usepackage{amsfonts}       
\usepackage{nicefrac}       
\usepackage{microtype}      
\usepackage{lipsum}
\usepackage{fancyhdr}       
\usepackage{graphicx}       
\graphicspath{{figures/}}     
\usepackage{amsmath}

\usepackage{indentfirst}
\usepackage{amsmath}
\usepackage{amssymb}

\title{ Complex-Valued Neural Networks - Theory and Analysis 
}

\author{
  Rayyan Abdalla \\
  \\
}

\begin{document}
\maketitle

\input{sections/abstract.tex}
\input{sections/introduction.tex}

\input{sections/Motivation}

\input{sections/Types_of_CVNNs}
\input{sections/Activation_functions}

\input{sections/Learning.tex}

\input{sections/Complex_Batch_Normalization}
\input{sections/complex_initialization}
\input{sections/CVNN_Implementation_Efforts}
\input{sections/future_research}

\bibliographystyle{unsrt}
\bibliography{references}

\end{document}

%% file: sections/abstract.tex
\begin{abstract}
  Complex-valued neural networks (CVNNs) have recently been successful in various pioneering areas which involve wave-typed information and frequency-domain processing. This work addresses different structures and classification of CVNNs. The theory behind complex activation functions, implications related to complex differentiability and special activations for CVNN output layers are presented. The work also discusses CVNN learning and optimization using gradient and non-gradient based algorithms. Complex Backpropagation utilizing complex chain rule is also explained in terms of Wirtinger calculus. Moreover, special modules for building CVNN models, such as complex batch normalization and complex random initialization are also discussed. The work also highlights libraries and software blocks proposed for CVNN implementations and discusses future directions. The objective of this work is to understand the dynamics and most recent developments of CVNNs.  
\end{abstract}

%% file: sections/Introduction.tex
\section{Introduction}
\label{sec:intro} 
Artificial neural networks have been tightly fused into various domains and applications fields. In many practical applications, including electromagnetics, signal and image processing, complex numbers occur by nature or design to represent information and frequency domain representations. This suggests that neural networks-based models utilized for such applications have parameters and variables in the complex domain. In general, the majority of widely deployed deep learning models are based on real-valued neural networks (RVNNs) due to less complexity and easier implementation associated with real parameters. Nonetheless, complex-valued neural networks (CVNNs), which are networks processing information with complex-valued parameters, have gained more interest and proven potential advantages over conventional RVNNs \cite{hirose2012complex,lee2022complex,bassey2021survey}. 

The prospects behind adopting CVNN can be viewed from biological, computational and application perspectives \cite{lee2022complex,bassey2021survey}. Complex-valued representations of neuron activity allows more versatile extension of brain physicality \cite{hirose2012complex}. Moreover, CVNNs are proved to perform more efficiently with better generalization for complex-valued tasks as well as nonlinear inseparable real-valued tasks. In terms of application fields, CVNNs recently have successful implementations in areas of image protection, image segmentation, image reconstruction, signal identification and estimation, telecommunications, speech processing, wind prediction, control systems and many other relevant domains \cite{lee2022complex}. This article explains the motivation when and why CVNNs can be more effective than RVNNs. Different neural networks structures that deal with complex-valued data are also discussed. Moreover, this article addresses CVNN distinctive features in details, including complex activation functions and CVNN learning algorithms, besides challenges and special considerations related to building CVNN based models. The article also lists most recent implementation efforts and highlights supporting software libraries for solving complex-valued problems. 

The work organization is as follows: section \ref{sec:Motiv} is CVNN motivation. Section \ref{sec:Types} provides proposed neural networks structures dealing with complex-valued representations. The complex activation functions are discussed in section \ref{sec:activation}. Learning algorithms and CVNN optimization is presented in section \ref{sec:learning}. Section \ref{sec:batch} explains implementation of batch normalization technique in the complex domain. Section \ref{complex weight} is complex weight initialization. Section \ref{sec:implement} summarizes implementation and software deployment attempts. Finally, future research directions and concluding remarks are drawn in section \ref{sec:research}.

%% file: sections/Motivation.tex
\section{Motivation of CVNN}
\label{sec:Motiv} 
Neural Networks based on Complex-valued representations have shown several advantages over their real-valued counterparts, though widely applied RVNNs manifest easier implementations regarding activation function design and learning algorithms. CVNNs have great potential in application domains that involve complex-valued data and signals \cite{lee2022complex,fuchs2021complex,scarnati2021complex,choi2018phase,fink2014predicting,tsuzuki2013approach}. Besides, CVNN models enable improved optimization, better generalization, and faster learning even for real-valued input representations \cite{barrachina2023theory,trabelsi2017deep} . In terms of biological viewpoint, complex-valued configuration of Neural Networks parameters proposes more versatile and biologically plausible neuron firing \cite{hirose2012complex,lee2022complex,bassey2021survey}. 

Several practical applications encode processed information using complex-valued data types. For example, image and signal processing involve frequency domain representations and expressions in terms of Transforms bases. Moreover, electromagnetic signals and wave-related applications, such as sensing, bio-informatics and telecommunications, employ amplitude-phase complex expressions to naturally represent signal energy and phase as progression in time. Neural networks utilized in such applications suggest that corresponding models use complex numbers to represent network inputs, outputs and learnable weights, i.e. using CVNNs. While many analyses propose that CVNNs can be implemented using double dimensional RVNNs, this may result in erroneous network behavior since complex number arithmetic is different from Real-valued operations and complex components are statistically correlated \cite{hirose2012generalization}. For wave-typed information, CVNNs are essentially more compatible than RVNNs since amplitude and phase are directly related to learning objective and real-world wave information can be fully extracted when training a CVNN rather than RVNN \cite{hirose2012complex}. 

In terms of computational motivation, CVNNs feature complex numbers arithmetic, particularly complex multiplication operation, for weights learning and input-output mapping. While complex multiplication yields amplitude attenuation and phase rotation, this results in an advantageous reduction in degrees of freedom and minimizing learnable parameter \cite{hirose2012complex} . In other words, RVNNs approach generally represents complex components independently either in vector or matrix form. That is, real-valued matrix operations on complex components increase the model parameters and dimensions, resulting in undesirable generalization characteristics and greater tendency for model overfitting \cite{hirose2012generalization}. On the other hand, CVNNs with complex weights utilize phase rotation dynamics to underpin learning task in one-dimensional complex domain, and matrix representation in the complex domain mimics rotation matrix with half of matrix entries already known. Therefore, CVNN handling of complex-valued problems results in reduced solution arbitrariness and enhanced generalization characteristics. The computational power of CVNNs is also revealed through the orthogonal property of the complex-valued neuron. This property refers to the concept of orthogonal decision boundary existing in the complex-valued neuron in which two hypersurfaces intersect and divide decision boundary into four regions. This high generalization ability enables solving special problems, such as the XOR problem, that cannot be solved by a single real-valued neuron \cite{nitta2003solving}. Furthermore, representations based on complex numbers have additional computational advantages for neural networks with architectures similar to associative memories, such as Deep residual networks (ResNets) and recurrent neural networks (RNNs). It has been shown that Holographic Reduced Representations, which use complex numbers, enable numerically efficient insertion and retrieval into associative memory in key-value pairs, increasing representational capacity and stability \cite{plate2003holographic}. In addition, orthogonal weight matrices have been proposed to mitigate the problem of vanishing or exploding gradients associated with long-term dependencies in network learning process. Unitary weights matrices, based on which Unitary RNNs were proposed, are generalization of the orthogonal matrices in the complex plane. Unitary RNNs uncover spectral representations by implementing discrete Fourier transform to learn unitary weight matrices \cite{arjovsky2016unitary}. That is, this enables richer representations by spanning parts of the matrix full set and preserve gradient over learning time.  

From biological perspective, using complex-valued weights proposes plausible neuronal formulation that allows constructing a richer, versatile deep network representation. In general, neurons convey information in the form of rhythmical spikes characterized by average firing rate and comparative timing of activity \cite{reichert2013neuronal}, which can be specified by amplitude and phase of a complex valued neuron. On the contrary, real-valued neuron representation considers only the firing rate and neglects the phase, which may affect neurons synchronization. With complex formulation, input neurons of similar phase are synchronous and add constructively, those with dissimilar phases add destructively and are terms asynchronous.  Therefore, inclusion of phase information is effective to synchronize inputs that the network propagates with time at a given layer, useful for controlling mechanisms of deep gating-based networks.

Given all these supportive evidences from computational, biological and signal processing perspectives, It is reasonable to apply complex numbers to neural network models.

%% file: sections/Types_of_CVNNs.tex
\section{Neural Networks for Complex-valued Data Representations}
\label{sec:Types}

\subsection{RVNN Implementation}

 There are two possible approaches for handling complex-valued inputs using conventional RVNNs. In both methods, the individual components of complex inputs are treated independently as real-valued numbers. Such RVNN variants can be described as follows \cite{hirose2012generalization}:
\subsubsection{Real-valued Bivariate Network Structure:}
 RVNN with double input terminals for real and imaginary parts of the input, and double output neurons to generate real and imaginary parts of output signal. Therefore, forward processing is performed in a bivariate manner by doubling the number of neurons of different network layers.

\subsubsection{Dual Real-valued Univariate Network Structure:}
This is a type of RVNN where real and imaginary parts of the input are processed separately. Many techniques can be applied to separate or mix the real and imaginary parts. That is, the neurons from the real-part network are not connected to the neurons from the imaginary-part network. Therefore, the respective networks will not double the number of terminals and output neurons. 
\\
\par
Handling complex input with real-valued networks increases network dimensions and learnable parameters \cite{lee2022complex,hirose2012complex}. Moreover, RVNN approaches result in phase distortion and decreased approximation accuracy as real-valued gradient does not reflect true complex-valued gradient \cite{lee2022complex}.  

\subsection{CVNN Implementation} \label{cvnn_types}
As CVNNs retain complex-valued parameters and variables, there are two types of CVNNs \cite{lee2022complex}:

\subsubsection{Split-CVNNs:}
Here, the input value splits into two real-valued components in the form of either rectangular coordinates (real and imaginary components) or polar coordinates (magnitude and phase components). The split components are fed to a network with complex valued weights and real-valued activation function \cite{barrachina2023theory,trabelsi2017deep}. 

\subsubsection{Fully Complex CVNNs:}
Such networks deal directly with complex-valued signals, where both weights and activation functions are in the complex domain \cite{kim2000fully,hirose1992proposal}.\\ 

\par
Unlike RVNN implementation, CVNN approaches overcome phase distortion problem and result in minimum generalization error for highly coherent signals \cite{hirose2012generalization}. Nonetheless, Fully complex CVNN are computationally complex as they require complex-differentiable (i.e. Holomorphic) activation function, whereas Split-CVNN are mostly-implemented CVNN since they make use of conventionally available non-holomorphic activation functions. 

%% file: sections/Activation_functions.tex
\section{Complex Activation Functions}
\label{sec:activation} 
Activation functions are required to introduced non-linear mapping to affine transforms in artificial neural networks. Given an input signal $x_i$, a connection weight $w_i$, an internal state $u = \sum_{i} x_i w_i$, activation function $f$, the output $y$ is given by:

\begin{equation}
    y = f(u) = f\left( \sum_{i} w_i x_i \right)
\end{equation}
In general, activation functions are chosen to be peicewise smooth to facilitate gradient computation during the learning process. Due to complex differentiability concerns, designing a suitable activation function for CVNN is one of the most challenging parts that hindered earlier CVNN developments \cite{hirose2012complex}. 

\subsection{Concerns Regarding Activation Function Differentiability in the Complex Domain} \label{concerns}

In terms of differentiability in the complex domain, functions are though to be \textit{holomorphic}, i.e. complex differentiable, or \textit{non-holomorphic}. For a complex function to be holomorphic, it must satisfy Cauchy-Riemann equations \cite{barrachina2023theory,trabelsi2017deep}. Designing complex activation functions conforms Liouville's theorem which implies that \textit{bounded} real-valued activation functions are non-holomorphic in the complex plane $\mathbb{C}$. Therefore,  it is restrictedly difficult to directly extend conventional real-valued activation functions in the complex domain. This is further elaborated in the following:

\subsubsection{Holomorphism and Cauchy-Riemann Equations}
Holomorphism, also termed analyticity, in a domain within the complex coordinate space refers to that a complex function is differentiable in the neighborhood of every point within the corresponding domain. A complex function that is differentiable in the whole complex plane $\mathbb{C}$ is called \textit{entire} function. That is, for a complex number $z = x+iy$ and a complex-valued function $f(z) = u(x,y)+iv(x,y)$ where $u$ and $v$ are real-valued functions, then $f$ is holomorphic in a complex domain if the derivative $f^{'}(z_0)$ exists at every point $z_0$ in that domain.
\begin{equation} \label{f_deriv}
    f^{'}(z_0) =  \lim_{\Delta z\to 0} \left[ \frac{\left( f(z_0) + \Delta z\right) - f(z_0)}{\Delta z}\right] = \lim_{\Delta x\to 0} \lim_{\Delta y\to 0} \left[ \frac{\Delta u(x_0,y_0) +i \Delta v(x_0,y_0)}{\Delta x + i \Delta y}\right]
\end{equation}

Since $\Delta z = \Delta x + i \Delta y$, $\Delta z$ can approach $0$ from multiple directions whether along real axis (i.e. $\Delta y = 0$), imaginary axis (i.e. $\Delta x = 0$ or in-between axes.However, for the function $f$ to be complex differentiable, $f^{'}(z_0)$ must give the same quantity regardless the direction of approach. Consider $\Delta z$ approaches $0$ along real axis or imaginary axis, respectively:
\begin{equation} \label{f_deriv_xy}
    f^{'}(z_0) =  \lim_{\Delta x\to 0} \left[ \frac{\Delta u(x_0,y_0) +i \Delta v(x_0,y_0)}{\Delta x + i 0}\right] = \lim_{\Delta y\to 0} \left[ \frac{\Delta u(x_0,y_0) +i \Delta v(x_0,y_0)}{ 0 + i \Delta y}\right]
\end{equation}
The above equation is equivalent to:
\begin{equation} \label{fz}
    \frac{\partial f}{\partial z} = \frac{\partial u}{\partial x} + i \frac{\partial v}{\partial x} = -i \frac{\partial u}{\partial y} +  \frac{\partial v}{\partial y} 
\end{equation}
Following equations \ref{f_deriv_xy} and \ref{fz}, the function $f$ is holomorphic if it satisfies the two equations:

\begin{equation} \label{cauch_reimann}
    \frac{\partial u}{\partial x} = \frac{\partial v}{\partial y}, \frac{\partial u}{\partial y} = - \frac{\partial v}{\partial x}
\end{equation}
The equations in \ref{cauch_reimann} are Cauchy-Riemann equations which represent a necessary condition for holomorphism. If the functions $u$ and $v$ have continuous first derivatives, then the equations \ref{cauch_reimann} are sufficient condition for a holomorphic $f$ \cite{trabelsi2017deep}.

In order to analyze complex neural networks dynamics and learning in the same manner as in real-valued neural networks, proposed nonlinear complex activation function must be analytic over whole $\mathbb{C}$ to avoid the problem of exploding or vanishing gradients during backpropagation .

\subsubsection{Liouville's Theorem} \label{liou}
In complex analysis, a function $f(z)$ is called \textit{bounded} if it satisfies: $|f(z)| < M, M \in \mathbb{R^+} $. Liouville's Theorem states that if a function $f(z)$ that is analytic, i.e. holomorphic at any point in $ \mathbb{C}$, and is bounded, it must be a \textit{Constant} function: $f(z) = c, c \in \mathbb{C} $. Equivalently, entire nonlinear functions, that are differentiable in whole plane $\mathbb{C}$, are unbounded functions, making them unsuitable as CVNN activation functions \cite{barrachina2023theory, bassey2021survey, lee2022complex}.

\subsection{CVNN Activations with Complex Partial Derivatives} \label{cvnn_par}
The complex differentiation concerns mentioned previously constrain CVNN activation function design by selecting one property, either bounded or analytic. By bounding the domain of the complex function and isolating singular points, one can propose an activation function that is holomorphic in the corresponding domain \cite{trabelsi2017deep}. Nevertheless, according to \cite{hirose2012complex}, complex activation functions can also be designed without paying attention to differentiability, instead, CVNN dynamics and self-organization can be constructed by means of partial derivatives with respect to complex components. An activation function shape can be determined with respect to input/output properties and processing purposes, meaning that non-holomorphic and complex linear activations are also possible. 

Non-holomorphic function differentiation can be done utilizing Wirtinger Calculus \cite{kreutz2009complex} to compute the gradient of complex components in part. In other words, complex backpropagation and CVNN optimization can be implemented using non-holomorphic complex activation functions by computing the partial derivatives with respect to real and imaginary parts. This therefore facilitates complex chain rule implementation for both real and complex functions. Further explanation is provided in the following:

\subsubsection{Wirtinger Calculus} \label{wircal}
Wirtinger calculus provides a generalized framework for computing complex function derivative, with the holomorphic function being a special case. For a complex function $f(z)$ of a complex variable $z = x + iy$, the partial derivative with respect to $z$ and the conjugate $\Bar{z}$ \cite{barrachina2023theory}:
\begin{equation} \label{wirtinger}
    \begin{split}
        \frac{\partial f}{\partial z} = \frac{1}{2} \left(  \frac{\partial f}{\partial x} - i  \frac{\partial f}{\partial y}\right) \\
     \frac{\partial f}{\partial \Bar{z}} = \frac{1}{2} \left(  \frac{\partial f}{\partial x} + i  \frac{\partial f}{\partial y}\right) 
    \end{split}
\end{equation}
From \ref{wirtinger},  the derivatives of $f$ with respect to real part $x$ and imaginary part $y$ are given by:
\begin{equation} \label{parts_deriv}
    \begin{split}
        \frac{\partial f}{\partial x} = \frac{\partial f}{\partial z} +  \frac{\partial f}{\partial \Bar{z}}  \\
        \frac{\partial f}{\partial y} = i \left(  \frac{\partial f}{\partial z} -  \frac{\partial f}{\partial \Bar{z}}  \right) 
    \end{split}
\end{equation}
Based on Wirtinger formulation and for a holomorphic function $f$ satisfying Cauchy-Riemann condition in \ref{cauch_reimann}, then: $ \frac{\partial f}{\partial \Bar{z}} = 0$, as proved in theorem 2.5 of \cite{barrachina2023theory}. This condition makes holomorphic functions more computationally efficient since gradient values are shared during backpropagation \cite{trabelsi2017deep}, as will be shown in later sections. 

\subsubsection{Complex Chain Rule} \label{chainrule}
Following Wirtinger Calculus framework, chain rule can be implemented for both real and complex functions with respect to complex variables. Analogous to multivariate chain rule with real values, as shown in theorem A.3 of \cite{barrachina2023theory}, given a complex variable $z = x+iy$, complex functions $h,g$ where $h$ is a function of $g$ is a function of $z$ such that $h(g(z))$. Theorem A.4 and A.5 of \cite{barrachina2023theory} state that the complex chain rule is given by:

\begin{equation}
    \begin{split}
         \frac{\partial h(g)}{\partial z} = \frac{\partial h}{\partial g} \frac{\partial g}{\partial z} + \frac{\partial h}{\partial \Bar{g}} \frac{\partial \Bar{g}}{\partial z} \\
         \frac{\partial h(g)}{\partial \Bar{z}} = \frac{\partial h}{\partial g} \frac{\partial g}{\partial \Bar{z}} + \frac{\partial h}{\partial \Bar{g}} \frac{\partial \Bar{g}}{\partial \Bar{z}}
    \end{split}
\end{equation}
And the complex chain rule with respect to real and imaginary parts of $z$:
\begin{equation}
    \begin{split}
         \frac{\partial g}{\partial z} = \frac{\partial g}{\partial x} \frac{\partial x}{\partial z} + \frac{\partial g}{\partial y} \frac{\partial y}{\partial z} \\
    \end{split}
\end{equation}

There is no assumption regarding whether the complex functions $h,g$ are holomorphic or not. Note that real valued functions are already included in the complex domain and follow the same chain rule with respect to complex variables. Mathematical proofs of the formulas are provided in  \cite{barrachina2023theory}.

\subsection{Types of CVNN Activation Functions}
The complex domain widens the possibilities to design a complex activation function. Taking into account all previously mentioned complex analysis, two categories of CVNN activation functions are proposed \cite{lee2022complex,bassey2021survey} , upon which CVNN types were defined as in \ref{cvnn_types}. First, Split activation functions, which deals with complex components independently. Second, Fully complex activation functions, in which complex components are treated as a single entity. 
\subsubsection{Split Activation Functions} \label{splitfunc}
The idea of split activation function was motivated by ability to differentiate non-holomorphic functions with respect to their complex component parts, as explained in \ref{cvnn_par}. Split activation functions are adopted from real-valued state-of-the-art deep architectures, that is, such CVNN activations apply real-valued conventional activation function to complex component parts independently and combine results to produce complex neuron output . Therefore, such activation function are bounded and not analytic. Based on complex coordinate system, i.e. cartesian coordinates or polar coordinates, which is imposed by CVNN application, split activation function are further divided into two types \cite{hirose2012complex,bassey2021survey,barrachina2023theory,lee2022complex}. For a complex variable $z = x+iy = |z| \exp(i \arg(z)) , z \in \mathbb{C}$ and $x,y \in \mathbb{R}$
\begin{enumerate}
    \item Type-A Split Activation Function: $f_A(z) = f_{Re}(x)+i f_{Im}(y)$
    \item Type-B Split Activation Function $f_B(z) = f_{r}(|z|)+i f_{\phi}(\arg(z))$
\end{enumerate}
Where $f_A, f_B$ are complex-valued functions,  and $f_{Re}, f_{Im}, f_r, f_{\phi}$ are real-valued activation functions of conventional types ( can be Sigmoid, tanh, Identity ...). In general, CVNNs employing Type-A activation functions select the same  real function for $f_{Re}$ and $f_{Im}$. Such networks work well with complex information having symmetry or special meaning on real and imaginary axes \cite{hirose2012complex}, and can also process real-valued data with half number of RVNN network parameters \cite{trabelsi2017deep,barrachina2023theory}. On the other hand, widely implemented Type-B activation functions squashes the magnitude and leave the phase unchanged, i.e. set $f_r$ as a conventional activation function while $f_{\phi}$ is identity. That is, networks with Type-B activation functions are adopted for wave-related applications and information of isotropic structure due to their direct correspondence to signal amplitude and phase\cite{hirose2012complex}.\par
Most of popular activation functions from RVNNs are extensible and easily implemented using Type-A and Type-B complex activation function, though the hyperbolic tangent function is already implemented in the complex domain and its transformation is not necessary \cite{barrachina2023theory}.

\subsubsection{Fully-Complex Activation Functions} \label{mvn}
Fully complex activations corresponds to using holomorphic functions that deal with complex variable as a single entity. Difficulty in fulfilling Liouville's theorem greatly restrict the use of fully complex activation function since the proposed function must be bounded and analytic in the desired complex domain \cite{trabelsi2017deep}.\par
Early works of fully complex activations involve the mapping of multi-valued neuron (MVN) \cite{aizenberg1973multivalued}. The MVN is a neural element with activation that maps the complex weighted input into an output lying on the unit circle. Specifically, for a complex weighted sum $z$ and activation function $ f$, the neuron output: $\omega^j = \exp(i2\pi j/k), j  = 0,1,...,k-1$, is one of $k$ roots of unity according to:
\begin{equation}
    \begin{split}
        f(z) = \exp (i2\pi j/k), \hspace{0.5in}  if \hspace{0.1in}   2\pi j/k \leq \arg (z) \leq 2 \pi (j+1)k
    \end{split}
\end{equation}
Moreover, MVN activation function can deal with continuous-valued inputs by making $k \to \infty$ \cite{bassey2021survey}, such that
\begin{equation}
    f(z) = \exp(i \arg (z)) = \frac{z}{|z|}
\end{equation}
Fully-complex MVN activation function can be also seen as a Type-B split activation function that sets the magnitude to unity and squashes the phase, as the case of "Phasor" networks \cite{hirose1992dynamics}. \par
Furthermore, elementary transcendal functions (ETFs) are proposed for fully-complex activations \cite{kim2000fully,georgiou1992complex}. These functions ( for example: sinh($z$), sin($z$), arcsin($z$), tan($z$), arctan($z$), arctanh($z$), arccos($z$) ) are holomorphic almost everywhere in the complex domain where singularities can be isolated or removed in the desired complex domain \cite{bassey2021survey,kim2003approximation}. However, training with ETFs is restrictedly limited due to lower generalization ability and slow convergence rate \cite{lee2022complex}. \par

There is still no agreed rule on which approach is best for CVNN activations. The only concern is that the selected activation function does not cause vanishing or exploding gradient during learning process. Nevertheless, other complex activation functions with structures slightly dissimilar to those described above were also proposed, such as complex adaptions of popular ReLU functions.

\subsection{Complex Rectified Linear Unit (ReLU)}
The RelU function, $ReLU(x) := max(x,0)$, is widely applied in conventional RVNNs due to proved cabaility of earning faster than equivalents with saturating neurons. Therefore, this motivated defining several variants of RelU function in the complex domain. As regards ReLU split activation, ReLU Type-B   activation function is not interesting since modulating the amplitude without the phase makes the activation converge to a linear function. Nevertheless, Type-A is successfully implemented and is therefore defined as Complex ReLU or C-ReLU \cite{barrachina2023theory,trabelsi2017deep}. \par
Another ReLU adaptation is also proposed and is called z-ReLU, which lets the output as the the input for points in the first quadrants in the complex domain \cite{guberman2016complex}: 
\begin{equation}
    z-ReLU =
    \begin{cases}
      z \hspace{0.5in} if \hspace{0.1in} 0 < \arg(z) < \pi/2\\
       0 \hspace{0.5in} if \hspace{0.1in} otherwise
    \end{cases}                  
\end{equation}
\par
Other ReLU adaptations attenuate the magnitude of the complex variable and preserve the phase. One function is the popular function modReLU \cite{arjovsky2016unitary}, which defines a radius $b$ in the complex plane along which the function output is $0$ and is defined as
\begin{equation}
    modReLU =
    \begin{cases}
      ReLU(|z|+b) \frac{z}{|z|} \hspace{0.5in} if \hspace{0.1in} |z| \geq b\\
       0                        \hspace{1.45in} if \hspace{0.1in} otherwise
    \end{cases}                  
\end{equation}
\par
another ReLU extension is the cardioid function \cite{virtue2017better}, which also leave the phase unchanged and modifies the magnitude based on the phase, and is defined as

\begin{equation}
    f_{cardioid} = \frac{(1+ \cos(\arg(z)))z}{2}
\end{equation}
All previously mentioned Complex ReLU variants were implemented and tested by \cite{barrachina2023theory,trabelsi2017deep}

\subsection{Output Layer Activation Function} \label{opact}
The nature of the final output neuron depends on the CVNN objective. That is, CVNN output may be required in the real domain $\mathbb{R}$. For Instance, in classification tasks, data labels may be provided as real-valued integers. Further, CVNNs may be adopted for solving linearly non-separable real-valued problems that cannot be solved by RVNNs or for computational efficiency \cite{lee2022complex}. Therefore, one proposed solution is to leave the complex output activation unchanged and cast labels with the transformation: $\text{for label } c \in \mathbb{R}, c \to c + ic$ \cite{barrachina2023theory}. \par
Another solution implies that activation functions specific for output layer are required to map complex input to real output. Thus, assuming $z = x +iy = |z|\exp(i\arg(z))$, input to output layer activation function $f$, conversion to real domain is done by \cite{lee2022complex,nitta2003solving}
\begin{equation}
         f(z) = |z|   \\
\end{equation}
or
\begin{equation}
    f(z) = (x-y)^2
\end{equation}
\par
Another possible mapping function is the \textit{softmax} function which maps integers to the interval $[0,1]$. the complex components of $z$ can be processed with real-valued operations ( such as averaging, multiplication or addition ) and then fed to the softmax activation function. The softmax function can also be applied to $z$ components independently and resultant values are then combined with real-valued operation. More details are found in \cite{barrachina2023theory}.

%% file: sections/Learning.tex
\section{Learning and Optimization in CVNN}
\label{sec:learning}
Neural network learning is the process of attractively tuning network weights in order to optimize a certain learning objective such as minimizing a loss function. The target is to achieve optimal set of weights that generalizes best to out-of-sample data. For CVNNs, the complex weights are tuned to minimize a real-valued loss function, as the minimum of two complex numbers is not feasible. There are many options to specify CVNN minimization criterion depending on network application and relevance of complex components. 

\subsection{Complex Compatible Loss Function}
In the context of supervised learning, the ground truth and desired network output is already known. For complex-valued regression tasks, given the desired and predicted network output, $d \in \mathbb{C}$ and  $o \in \mathbb{C}$, respectively, the error function is naturally given by \cite{lee2022complex,bassey2021survey}

\begin{equation} \label{E}
    E = \frac{1}{2} e.\Bar{e} = \frac{1}{2} |e|^2, \hspace{0.5in} \text{where}\hspace{0.2in} e = d - o \in \mathbb{C} 
\end{equation}
The regression loss of $N$ output neuron is then defined as
\begin{equation} \label{loss}
    \mathcal{L} = \sum_{k=1}^{N} E_k = \frac{1}{2} \sum_{k=1}^{N} |e_k|^2
\end{equation}
The above expression \ref{E} only represents the error in amplitude. However, for complex valued signals, the phase also carries important information. Therefore, another error function based on logarithm minimization was proposed \cite{lee2022complex}
\begin{equation}
    \begin{split}
        e_{log} = \left( log(o_k) - log(d_k)\right) \overline{\left( log(o_k) - log(d_k)\right)}\\
        E_{log} = \frac{1}{2} |e_{log}|^2 = \frac{1}{2} \left[ \left( log \frac{r_d}{r_o}\right)^2 + \left( \phi_d - \phi_o\right)^2 \right]
    \end{split}
\end{equation}
Where $r_d$, $r_o$ and $\phi_d$, $\phi_d$ are the magnitudes and phases of desired and predicted outputs, respectively, and the logarithmic loss function is computed as a summation of $E_log$ of output neurons. 
\par
For classification and semantic segmentation tasks, the loss function does not usually support complex-valued output, such as cross entropy function. One possible solution is to map complex output to the real domain using output activation function, as described in \ref{opact}. Another solution was proposed for evaluating categorical cross-entropy loss function for complex-valued outputs \cite{cao2019pixel}. This solution compares the real and imaginary parts of the predicted output independently with labels and computes the loss function using the equation
\begin{equation}
    \mathcal{L}_{ACE} = \frac{1}{2} \left( \mathcal{L}_{CE}(\Re{(o))},d) + \mathcal{L}_{CE}(\Im{(o)},d)\right)
\end{equation}
Where ${L}_{ACE}$ is the proposed average cross-entropy loss function for complex values and ${L}_{CE}$ is the conventional cross-entropy function utilized for classification tasks in RVNNs. All of the above loss functions were implemented by \cite{barrachina2023theory}.

\subsection{CVNN Learning Approaches}
The learning algorithm of conventional real-valued neural networks, which is backpropagation with gradient decent, requires optimizing the network using gradient or any similar partial-derivative technique. CVNN training, on the other hand, can be done using two possible approaches. One approach follows the same method of RVNN training with slight modifications to implement gradient decent and backpropagation in complex domain. The second approach implements error backpropagation without gradient decent and update weights analytically for each training sample \cite{bassey2021survey}. 
\subsubsection{Gradient-based Learning}
Similar to RVNN learning process, CVNN gradient-based learning calculates the error in the forward pass and then the error is backpropagated to each neuron of the network and weights are adjusted in backward pass. Adjusting CVNN weights with gradient decent requires evaluating derivative of the real-valued loss function with respect to complex-valued network weights and thus, applying chain rule. Due to complex differentiability concerns illustrated in \ref{concerns}, standard complex differentiation only exists for holomorphic functions. However, CVNN utilize non-holomorphic function in two ways: 
\begin{itemize}
    \item Real-valued loss function is non-holomorphic according to Liouville's theorem \ref{liou}.
    \item Split-activation functions \ref{splitfunc}, extended from RVNN implementation, are non-holomorphic.
\end{itemize}

However, the Wirtinger calculus, discussed in \ref{wircal}, is utilized to derive the complex gradient for both holomorphic and non-holomorphic functions and generalize the chain rule, as shown in \ref{chainrule}, to successfully implement CVNNs \cite{bassey2021survey,barrachina2023theory,lee2022complex}.\\

\begin{figure}[!htb]
    \centering
    \includegraphics[scale=0.7] {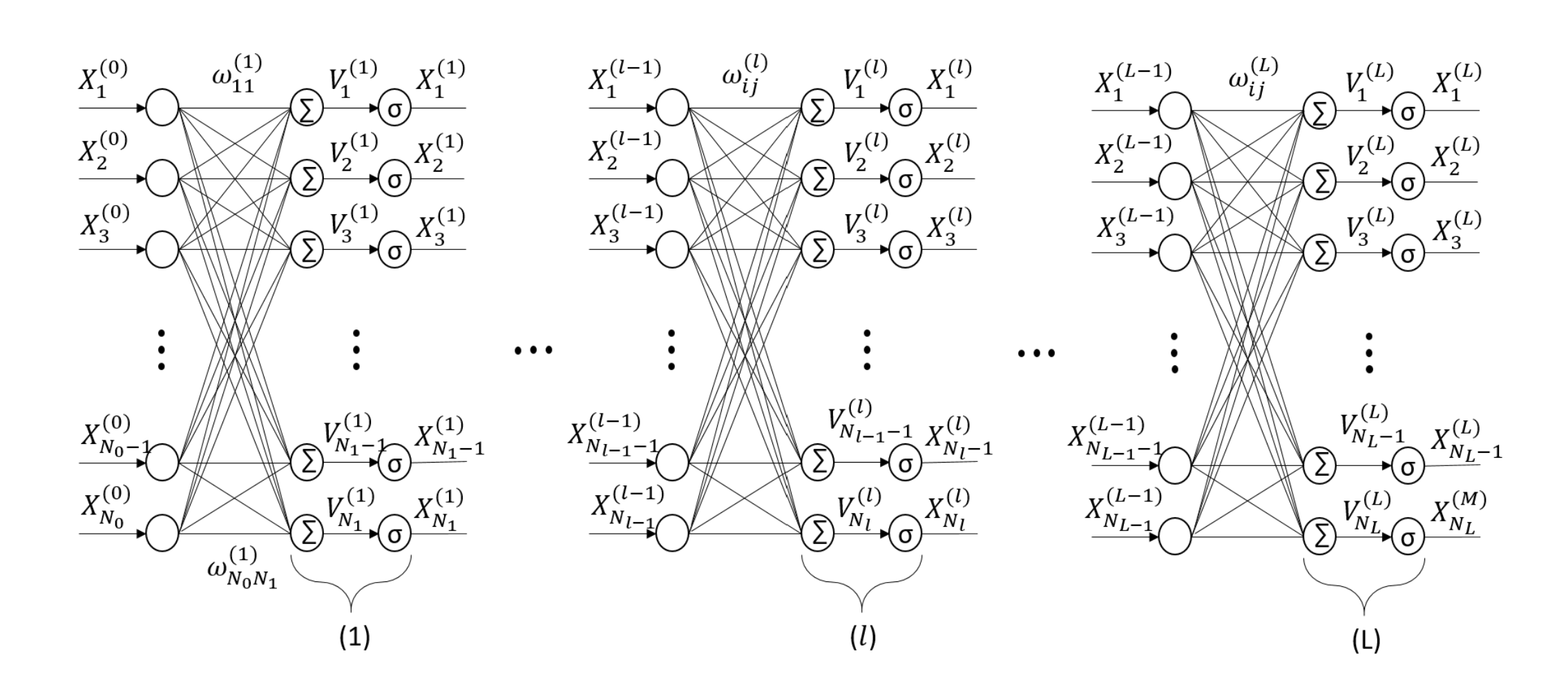}
    \caption{Feed-forward Neural Network Diagram}
    \label{fig:ffnn}
\end{figure}

For further analysis of CVNN dynamics, consider the multi-layer feedforward neural network shown in Figure \ref{fig:ffnn} from \cite{barrachina2023theory}. The network has $L$ layers indexed $0 \leq l \leq L$ where $l = 0$ corresponds to the input layer. Each layer $l$ has $N_l$ neurons indexed $1 \leq n \leq N_l$. The parameter $w^{(l)}_{nm}$ is the weight of the $n^{th}$ neuron of layer $l-1$ connected to the $m^{th}$ of layer $l$. The symbol $\sigma$ indicates the complex activation function. The output of layer $l$ and input to layer $l+1$ is $X^{(l)}_{n} = \sigma \left( V^{(l)}_{n} \right)$ where $V^{(l)}_{n} = \sum_{m=1}^{N_{l-1}} w^{(l)}_{nm} X^{(l-1)}_{m}$. The error function of neuron $n$ of output layer is denoted $E_n$ and the loss function is given by $\mathcal{L} = \sum_{n=1}^{N_L} E_n$

\paragraph{Complex Backpropagation}\hspace{0pt} \\
\par
The complex gradient of a complex-valued function $f$, relative to complex variable $z = x+iy$ and $f$ is differentiable with respect to real and imaginary parts, is given by \cite{amin2011wirtinger,leung1991complex,li2008complex}:
\begin{equation}
    \nabla_z f(z, \Bar{z}) = \frac{\partial f}{\partial x} + i \frac{\partial z}{\partial y}
\end{equation}
Which is, according to Wirtinger calculus \ref{wircal}, is evaluated as \cite{barrachina2023theory}
\begin{equation}
    \nabla_z f(z, \Bar{z}) = \overline{ \left( \frac{\partial f}{\partial z} + \frac{\partial f}{\partial \Bar{z}} \right) } = 2 \frac{\partial \Re(f)}{\partial \Bar{z}}
\end{equation}
For CVNN gradient decent implementation, the loss function is real valued, therefore it is natural to define gradient $\nabla_z f = 2 \frac{\partial f}{\partial \Bar{z}}$.

Recall that in RVNN learning, starting at the output layer, the the partial derivative of the loss function with respect to a single network weight is found with chain rule as follows
\begin{equation}
    \frac{\partial \mathcal{L}}{\partial w^{(l)}_{nm}} = \sum_{n=1}^{N_L}  \frac{\partial E_n}{\partial X^{(L)}_{n}}  \frac{\partial X^{(L)}_{n}}{\partial V^{(L)}_{n}}  \frac{\partial  V^{(L)}_{n}}{\partial w^{(l)}_{nm}}
\end{equation}
and the derivative components are evaluated by passing the network backwards with the chain rule. In case of CVNN learning, the chain rule is changed using \ref{chainrule} and the derivative of error function with respect to complex neuron weight is

\begin{equation} \label{pareparw}
    \frac{\partial E}{\partial w} = \frac{\partial E}{\partial X} \frac{\partial X}{\partial V} \frac{\partial V}{\partial w} + \frac{\partial E}{\partial X} \frac{\partial X}{\partial \Bar{V}} \frac{\partial \Bar{V}}{\partial w} + \frac{\partial E}{\partial \Bar{X}} \frac{\partial \Bar{X}}{\partial V} \frac{\partial V}{\partial w} + \frac{\partial E}{\partial \Bar{X}} \frac{\partial \Bar{X}}{\partial \Bar{V}} \frac{\partial \Bar{V}}{\partial w}
\end{equation}

Since $ E $ is real-valued, the conjugate rule applies as 
\begin{equation} \label{conjugate}
    \frac{\partial E}{\partial \Bar{X}} = \overline{\left( \frac{\partial E}{\partial X} \right) } \hspace{0.1in} , \hspace{0.1in} \frac{\partial \Bar{X}}{\partial \Bar{V}} = \overline{\left( \frac{\partial X}{\partial V} \right) } \hspace{0.1in} , \hspace{0.1in} \frac{\partial X}{\partial \Bar{V}} = \overline{\left( \frac{\partial \Bar{X}}{\partial V} \right) } 
\end{equation}
As explained in \cite{barrachina2023theory}, for $h \leq l-2$, the formulas for obtaining complex derivatives can be given by
\begin{equation} \label{v/w}
    \frac{\partial V^{(l)}_n}{\partial w^{(h)}_{jk}} = 
    \begin{cases}
        X^{(l-1)}_j & h = l \\
        w^{(l)}_{jk} \frac{\partial X^{(l-1)}_j}{\partial V^{(l-1)}_{j}}   \frac{\partial V^{(l-1)}_{j}}{\partial w^{(l-1)}_{jk}} & h = l-1 \\
        \sum_{i=1}^{N_{l-1}}  w^{(l)}_{ik} \left[  \frac{\partial X^{(l-1)}_i}{\partial V^{(l-1)}_{i}}   \frac{\partial V^{(l-1)}_{i}}{\partial w^{(l-1)}_{jk}}    +   \frac{\partial X^{(l-1)}_i}{\partial \Bar{V}^{(l-1)}_{i}}   \frac{\partial \Bar{V}^{(l-1)}_{i}}{\partial w^{(l-1)}_{jk}}   \right] & h \leq l-2
    \end{cases}
\end{equation}
And using conjugate properties

\begin{equation} \label{vc/w}
    \frac{\partial \Bar{V}^{(l)}_n}{\partial w^{(h)}_{jk}} = 
    \begin{cases}
        0 & h = l \\
        \Bar{w}^{(l)}_{jk} \frac{\partial \Bar{X}^{(l-1)}_j}{\partial V^{(l-1)}_{j}}   \frac{\partial V^{(l-1)}_{j}}{\partial w^{(l-1)}_{jk}} & h = l-1 \\
        \sum_{i=1}^{N_{l-1}} \Bar{w}^{(l)}_{ik} \left[  \frac{\partial \Bar{X}^{(l-1)}_i}{\partial V^{(l-1)}_{i}}   \frac{\partial V^{(l-1)}_{i}}{\partial w^{(l-1)}_{jk}}    +   \frac{\partial \Bar{X}^{(l-1)}_i}{\partial \Bar{V}^{(l-1)}_{i}}   \frac{\partial \Bar{V}^{(l-1)}_{i}}{\partial w^{(l-1)}_{jk}}   \right]  & h \leq l-2
    \end{cases}
\end{equation}
Using the fact that $\frac{\partial \Bar{V}^{(l-1)}_j}{\partial w^{(l-1)}_{jk}} = 0$, then equation \ref{pareparw} can be re-written as
\begin{equation} \label{pareparwnew}
   \begin{split}
        \frac{\partial E_n}{\partial w^{(l)}_{ji}} = \frac{\partial E_n}{\partial  X^{(l)}_{i}} \frac{\partial X^{(l)}_{i}}{\partial V^{(l)}_{i}} \frac{\partial V^{(l)}_{i}}{\partial  w^{(l)}_{ji}} + \frac{\partial E_n}{\partial \Bar{X}^{(l)}_{i}} \frac{\partial \Bar{X}^{(l)}_{i}}{\partial V^{(l)}_{i}} \frac{\partial V^{(l)}_{i}}{\partial  w^{(l)}_{ji}} \\
        = \frac{\partial E_n}{\partial  X^{(l)}_{i}} \frac{\partial X^{(l)}_{i}}{\partial V^{(l)}_{i}} X^{(l-1)}_{j} + \frac{\partial E_n}{\partial \Bar{X}^{(l)}_{i}} \frac{\partial \Bar{X}^{(l)}_{i}}{\partial V^{(l)}_{i}} X^{(l-1)}_{j}
   \end{split}
\end{equation}
Analyzing the error derivative with respect to layer input, the term $E_n/\partial X^{(l)}_i$ becomes
\begin{equation} \label{e/x}
    \frac{\partial E_n}{\partial X^{(l)}_i} = 
    \begin{cases}
        \frac{\partial E_n}{\partial X^{(L)}_n} & l = L\\

        \frac{\partial E_n}{\partial  X^{(L)}_{n}} \frac{\partial X^{(L)}_{n}}{\partial V^{(L)}_{n}} w^{(L)}_{in} + \frac{\partial E_n}{\partial \Bar{X}^{(l)}_{i}} \frac{\partial \Bar{X}^{(l)}_{i}}{\partial V^{(L)}_{n}} w^{(L)}_{in}   & l = L-1 \\

        \sum_{k=1}^{N_{l+1}} \left[ \frac{\partial E_n}{\partial  X^{(l+1)}_{k}}  \frac{\partial X^{(l+1)}_{k}}{\partial V^{(l+1)}_{k}} w^{(l+1)}_{ik} + \frac{\partial E_n}{\partial \Bar{X}^{(l+1)}_{k}} \frac{\partial \Bar{X}^{(l+1)}_{k}}{\partial V^{(l+1)}_{k}} w^{(l+1)}_{ik} \right] & l \leq L-2
    \end{cases}
\end{equation}
That is, $\partial E_n/\partial \Bar{X}^{(l)}_i$ can be obtained using the conjugate rule of \ref{conjugate}.
With equations \ref{v/w},\ref{vc/w} and \ref{e/x}, the backpropagation algorithm is fully defined in the complex domain.

\subsubsection{Non-gradient-based Learning}
This learning process of CVNN is based on MVN neural element with fully-complex phasor activation function, discussed in \ref{mvn}. Unlike the gradient-based approach, the MVN learning is derivative free and CVNN weights are updated with the error correction rule explained in \cite{aizenberg1973multivalued}. Upon determining the neuron error, the neural learning is done through a simple movement along the unit circle, i.e. only neuron phase is updated analytically with no gradient involved \cite{bassey2021survey}

Considering the same feed-forward network architecture of Figure \ref{fig:ffnn}, the network weights are initialized and the forward processing is evaluated. The complex network error is then estimated and is backpropagated starting from the output layer and the neuron error in each layer is used to update the weight of the corresponding neuron. Assume the error of neuron $m$ in the layer $l$ is $e_m^{(l)}$, the weight correction formula for hidden and output layer is 

\begin{equation}
    \Tilde{w}_{nm}^{(l)} = w_{nm}^{(l)} + \frac{C_m^{(l)}}{N_{l-1}+1} e_m^{(l)} \Bar{\Tilde{X}}^{l-1}_n
\end{equation}
And for the input layer, i.e. $l=0$
\begin{equation}
    \Tilde{w}_{nm}^{(0)} = w_{nm}^{(0)} + \frac{C_m^{(0)}}{n+1} e_m^{(0)} X^{0}_n
\end{equation}

Where $C_m^{l}$ is the learning rate for the $m^{th}$ of layer $l$. The learning process continues until a termination condition is met, which is generally average error magnitude of each layer falls under a threshold, sufficient to terminate the weights update \cite{bassey2021survey}.

%% file: sections/Complex_Batch_Normalization.tex
\section{Complex Batch Normalization} \label{batch normalization}
\label{sec:batch}

Batch normalization technique was originally proposed for RVNNs to accelerate learning and optimize the model \cite{ioffe2015batch}. As for CVNNs, special formulation is required to apply batch normalization in the complex domain. Standardizing an array of complex numbers such that their mean is 0 and variance is 1 cannot guarantee that resultant real and imaginary components are circularly distributed. In order to obtain equal variance in both components, the normalization problem can be assumed as whitening 2D vectors such that: $z = a+ib \in \mathbb{C} \to \textit{\textbf{x}} = (a,b) \in \mathbb{R}^2 $ \cite{trabelsi2017deep, barrachina2023theory}. That is, the complex batch normalization steps are explained as follows:

For a complex variable $\textit{\textbf{x}}$ with estimated mean $\Tilde{\pmb{\mu_x}} \in \mathbb{R}^2$ and a covariance matrix $ \Tilde{\pmb{\mathit{V}}}\in \mathbb{R}^{2\times2}$, the normalized output $\hat{\textit{\textbf{x}}}$ is given by
\begin{equation} \label{normout}
    \hat{\textit{\textbf{x}}} = \Tilde{\pmb{\mathit{V}}}^{-\frac{1}{2}} (\textit{\textbf{x}} - \Tilde{\pmb{\mu_x}} )
\end{equation}
Where 
\begin{equation}
    \Tilde{\pmb{\mathit{V}}} =
    \begin{bmatrix}
        V_{rr} &  V_{ri}\\
        V_{ir} & V_{ii} 
    \end{bmatrix} 
    = 
    \begin{bmatrix}
        \text{Cov} \left( \Re{(\textit{\textbf{x}})},\Re{(\textit{\textbf{x}})} \right) &  \text{Cov}\left( \Re{(\textit{\textbf{x}})},\Im{(\textit{\textbf{x}})} \right) \\
        \text{Cov}\left( \Im{(\textit{\textbf{x}})},\Re{(\textit{\textbf{x}})} \right) & \text{Cov}\left( \Im{(\textit{\textbf{x}})},\Im{(\textit{\textbf{x}})} \right) 
    \end{bmatrix} 
\end{equation}
This normalization procedure allows decorrelation of the real and imaginary components to avoid co-adaptation and possible overfitting \cite{trabelsi2017deep}. 

As for batch normalization layer initialization, a moving variance variable $\pmb{\mathit{V}}^{'}$ is initialized to $\mathbf{I}/\sqrt{2}$ and a moving mean $\pmb{\mu_x^{'}}$ is initialized to $\mathbf{0}$, to be used later by the moving average rule. 

During the training phase, $\Tilde{\pmb{\mu_x}}$ and $\Tilde{\pmb{\mathit{V}}}$ are estimated for each input batch along the desired principal component and then the normalization output is estimated using \ref{normout}.

Similar to batch normalization in RVNNs, the normalized output can be shifted and scaled using two learnable parameters, $\pmb{\beta}$ and $\pmb{\gamma}$. The shift parameter $\pmb{\beta} = \left( \Re{(\pmb{\beta})} \right) , \left( \Im{(\pmb{\beta})} \right)^T $ has two learnable components and is initialized to $\mathbf{0}$. The scale parameter $\pmb{\gamma}$  is a $2\times2$ matrix has three components to learn since off-diagonal elements are equal, and is initialized to $\mathbf{I}/\sqrt{2}$. Therefore, the complex batch normalization output upon scaling and shifting is given by
\begin{equation}
    BN(\hat{\textit{\textbf{x}}}) = \pmb{\gamma} \hat{\textit{\textbf{x}}}+\pmb{\beta}
\end{equation}

In addition, in the moving average step, batch normalization layer also use running averages with momentum $\alpha$ to keep an estimate of  mean and variance during the training phase. The moving mean and moving variance are iteratively updated as follows:
\begin{equation}
    \pmb{\mu_x^{'}}_{n+1} = \alpha \pmb{\mu_x^{'}}_{n} + (1-\alpha)\Tilde{\pmb{\mu_x}}_{n}
\end{equation}
\begin{equation}
    \pmb{V^{'}}_{n+1} = \alpha \pmb{V^{'}}_{n} + (1-\alpha)\Tilde{\pmb{V}}_{n}
\end{equation}
That is, in the inference phase, no batch statistics are computed and the layer output is computed directly using \ref{normout} but with $\pmb{\mu_x^{'}}$ and $\pmb{V^{'}}$.

%% file: sections/complex_initialization.tex
\section{Complex Weight Initialization} \label{complex weight}
In neural networks learning, proper weight initialization is essential to avoid the risk of vanishing or exploding gradients. A complex weight initializer requires defining suitable variance for complex weights components. Two formulations, based on either polar or rectangular complex representations, were proposed for CVNN weight initialization \cite{barrachina2023theory,trabelsi2017deep}. Both formulations estimate the variance of complex parameters following a criterion analyzed by \cite{glorot2010understanding}, which ensures almost constant variance for layers input, output and their gradients. 

\subsection{Complex Initialization from Polar Representation Perspective}
Consider a complex weight $w = |w|\exp{(-i \phi)} = \Re{(w)}+\Im{(w)}$, where $|w|$ and $\phi$ are the weight magnitude and phase, respectively. The variance of $w$ is given by

\begin{equation} \label{varw}
    \text{Var}(w) = \text{E} \left[ w \Bar{w}\right] - \left( \text{E} [w] \right)^2 = \text{E} \left[ |w|^2\right] - \left( \text{E} [w] \right)^2
\end{equation}
Assuming $w$ follows circular symmetric distribution, i.e. $\text{E} [w] = 0$, then \ref{varw} reduces to $\text{Var}(w) = \text{E} \left[ |w|^2\right] $. Moreover, symmetric distribution of $w$ implies normally distributed real and imaginary components, that is, the magnitude $|w|$ follows Rayleigh distribution (Chi-Squared distribution with two degrees of freedom). Thus, it is reasonable to find $Var(w)$ in terms of $Var(|w|)$ as follows
\begin{equation}
    \text{Var} \left( |w| \right) = \text{E}\left[ |w|^2\right] - (\text{E}\left[ |w|\right])^2 = \text{Var}(w) - ( \text{E}\left[ |w|\right])^2
\end{equation}
For a Rayleigh distribution with parameter $\sigma_r$, the mean and variance of $|w|$ are given by
\begin{equation}
    \text{E}\left[ |w|\right] = \sigma_r \sqrt{\frac{\pi}{2}}\hspace{0.2in}, \hspace{0.2in}\text{Var} \left( |w| \right) = \frac{4-\pi}{2} \sigma^2
\end{equation}
Therefore, the variance of $w$ becomes
\begin{equation}
    \text{Var}(w) = \text{Var} \left( |w| \right) + ( \text{E}\left[ |w|\right])^2 = 2 \sigma_r^{2}
\end{equation}
According to \cite{glorot2010understanding}, a good initialization compromise that maintains almost constant variance of layers weights during training is that $\text{Var}(w) = 2/\left(N_{in}+N_{out}\right)$, where $N_{in}$ and $N_{out}$ are the number of input and out weights corresponding to each layer. In order to satisfy this criterion, initializing weights magnitude in polar form follows Rayleigh distribution with parameter $\sigma_r = 1/\sqrt{\left(N_{in}+N_{out}\right)} $. The phase of the complex weights is initialized following Uniform distribution between $0$ and $\pi$. Subsequently, multiplying the randomly generated magnitude with the generated phasor gives initialized weight parameter \cite{barrachina2023theory,trabelsi2017deep}.

\subsection{Complex Initialization from Rectangular Representation Perspective}
According to the work of \cite{barrachina2023theory}, the real and imaginary components of complex weights can be initialized independently with a suitable constant varaince for both components. Considering the network architecture of figure \ref{fig:ffnn}, under the assumption of constant variance for all input features, statistically centered weights and constant weights variance related to a single layer, the necessary condition to keep constant information flow in forward and direction is
\begin{equation}
    \text{Var}X_n^{(l)} =  \text{Var}X_n^{(l')} ,  \forall 1 \leq l \leq l' \leq N
\end{equation}
And the necessary condition for constant flow in the backward direction is
\begin{equation}
     \text{Var} \left[ \frac{\partial \mathcal{L}}{\partial V_n^{(l)} }\right] =\text{Var} \left[ \frac{\partial \mathcal{L}}{\partial V_n^{(l')} }\right]  ,  \forall 1 \leq l \leq l' \leq N
\end{equation}
The above conditions can only be jointly met if all network layers have the same width so that the variance is given by $\text{Var} (w) = 1/N$ \cite{barrachina2023theory}. Therefore, the criterion proposed by \cite{glorot2010understanding} provides a good trade-off for $\text{Var}(w) = 2/\left(N_{in}+N_{out}\right)$. That is, the weight initialzation can follow a uniform distribution with this variance such that
\begin{equation}
    w \sim U\left[- \frac{\sqrt{6}}{\sqrt{\left(N_{in}+N_{out}\right)}}, \frac{\sqrt{6}}{\sqrt{\left(N_{in}+N_{out}\right)}},\right]
\end{equation}
Since real and imaginary parts are statistically independent, the weight variance can be expressed as
\begin{equation}
    \text{Var}(w) = \text{Var}\left( \Re{(w)}\right) + \text{Var}\left( \Im{(w)}\right)
\end{equation}
Thus, initializing weights components is done by
\begin{equation}
    \Re{(w)} = \Im{(w)} \sim U\left[- \frac{\sqrt{3}}{\sqrt{\left(N_{in}+N_{out}\right)}}, \frac{\sqrt{3}}{\sqrt{\left(N_{in}+N_{out}\right)}},\right]
\end{equation}

%% file: sections/CVNN_Implementation_Efforts.tex
\section{CVNN Implementation Efforts}
\label{sec:implement}
The implementations of complex-valued deep neural networks have always been related to specific structures and application scenarios. That is, CVNNs mathematical formulation is difficult to implement and yet popular Python libraries for implementing neural networks, which are \textit{PyTorch} and \textit{TensorFlow}, do not fully support CVNN model creation . In fact,  \textit{TensorFlow} supports the use of complex data types for automatic differentiation algorithm, it has been verified by \cite{barrachina2023theory} that \textit{TensorFlow} calculates complex gradients using Wirtinger Calculus. On the other hand, \textit{PyTorch} added complex data types as BETA starting with version 1.6 and also supported complex differentiation with Wirtinger Calculus. Later on with version 1.12, complex convolution functionality was added by \textit{PyTorch}. These are most recent CVNN developments by PyTorch and TensorFlow, which indicates tendency towards CVNN support.  

There are several libraries and code repositories available to develop CVNN models. The very first and probably most popular CVNN implementation is the GitHub code \footnote{\url{https://github.com/ChihebTrabelsi/deep_complex_networks}} which reproduce experiments presented by the paper \cite{trabelsi2017deep}. This work was originally proposed to solve real-valued problems using complex-valued deep neural networks. The authors provided analysis and building blocks required to design complex-valued neural networks structures, particularly complex convolutional residual networks and recurrent neural networks, targeting richer representational capacity and more efficient memory retrieval mechanisms. The complex components and algorithms presented are complex batch normalization in section \ref{batch normalization}, complex weights initialization in section \ref{complex weight} and complex convolutions. Their model was designed to process complex-valued information with double dimensional real-valued data types and that imaginary parts are learnt during training for real-valued data. Nonetheless, the implemented code (662 stars and 272 forks ) uses \textit{Theano} back-end, which is no longer maintained. Motivated by this, many other implementation attempts, based on either \textit{TensorFlow Keras} or \textit{PyTorch}, reproduced the work of \cite{trabelsi2017deep}. Such attempts include the library with GitHub code repository \cite{dramsch2019complex} (122 stars) which uses \textit{Keras} with \textit{TensorFlow} back-end. Another re-implementation attempt is the library with GitHub code \footnote{\url{https://github.com/wavefrontshaping/complexPyTorch/tree/master}} (459 stars), which uses \textit{PyTorch} v1.7 with fully complex tensors.  

Furthermore, another library that enables implementing comprehensive CVNN models was proposed by \cite{barrachina2023theory}. This library natively supports complex number data-types and supports a wide range of split-type and fully complex activation functions. Moreover, the library also supports special complex components proposed earlier by \cite{trabelsi2017deep}, which are complex batch normalization and complex initialization.  The code uses \textit{TensorFlow} as a back-end and was published in Zenodo \footnote{\url{https://doi.org/10.5281/zenodo.7303587}} and GitHub (107 stars) \cite{j_agustin_barrachina_2022_7303587}. The library also support implementing RVNN models and gives results comparable to those obtained by Conventional \textit{TensorFlow} based models.

Other codes repositories involving complex-valued neural networks are also available. However, such repositories are related to specific applications and neural networks variants with models that are less generalized. For this reason, the mathematical analysis of this article relies on the theory and implementations by \cite{barrachina2023theory} and \cite{trabelsi2017deep}. 

%% file: sections/future_research.tex
\section{Conclusion and Possible Research Directions} \label{sec:research}

The research and advancements on CVNNs have not been very active compared to RVNNs. This is due to several challenges and difficulties associated with CVNN architectures and learning mechanisms. This section highlights future research prospects and potential directions towards establishing effective CVNN models.

One of the most obvious limitations in CVNN training is the design of complex activation function. Due to Liouville's theorem, complex activations compromise one crucial property, either analytic or bounded. Moreover, split-type activation functions in polar form does not account for phase information during training. Therefore, an active research area is the design of fully-complex nonlinear activation function that is differentiable over widely applicable complex plane. Another proposed research direction is expressing CVNN learning rate in the complex domain. The complex-valued learning rate extends the gradient search area in the complex domain and allows escaping saddle points. For this reason, selection of adaptive generalized complex-valued learning rate can be beneficial for CVNNs. As regards learning algorithms, the difficulties associated with complex differentiability suggests exploring learning algorithms that does not involve gradient computations \cite{lee2022complex}. Another reason for slow development of CVNN models is the lack of public libraries for CVNN training. Even standard deep learning libraries, based TensorFlow or PyTorch, does not fully support CVNN implementation. Therefore, new deep learning libraries are required to optimize and train CVNN models. Moreover, more generalized complex building blocks, based on proposing statistical distributions for generating complex parameters such as complex random initialization and complex batch normalization, is also an active research topic. Hence, future studies are required to develop more consistent building blocks that are less sensitive to random complex parameters \cite{bassey2021survey}. Further, recalling the biological motivation behind CVNN systems, an interesting research area is the development of biologically plausible neuronal model that mimics the human neuronal information transmission. In other words, such novel biological CVNN systems may be beneficial for nonlinear real-world problems that favor controlling the neuron firing time and magnitude.

In conclusion, this work summarized the theory related to CVNN dynamics and corresponding implementations. Deep learning models based on CVNNs can be successfully applied following the aforementioned formulations and learning algorithms. This article is good start for understanding CVNN different structures and potential research directions.